\documentclass[11pt]{article}

\usepackage[margin=1in]{geometry}
\usepackage{microtype}
\usepackage{hyperref}
\usepackage{url}
\usepackage{booktabs}
\usepackage{graphicx}
\usepackage{multirow}
\usepackage{tabularx}
\usepackage{amsmath}
\usepackage{amssymb}
\usepackage{xcolor}
\usepackage{enumitem}
\usepackage{fvextra}
\usepackage{placeins}
\usepackage{tcolorbox}
\usepackage{tikz}
\usepackage{authblk}
\usetikzlibrary{shapes.geometric, arrows.meta, positioning, fit, backgrounds, calc}

\definecolor{darkblue}{rgb}{0, 0, 0.5}
\hypersetup{colorlinks=true, citecolor=darkblue, linkcolor=darkblue, urlcolor=darkblue}

\title{Measuring Representation Robustness in Large Language Models for Geometry}

\author[1]{Vedant Jawandhia}
\author[1]{Yash Sinha}
\author[3]{Murari Mandal}
\author[2]{Ankan Pal}
\author[1]{Dhruv Kumar}
\affil[1]{Department of Computer Science and Information Systems, BITS Pilani \\\texttt{\{f20220627, dhruv.kumar, yash.sinha\}@pilani.bits-pilani.ac.in}}
\affil[2]{Department of Mathematics, BITS Pilani \\ \texttt{ankan.pal@pilani.bits-pilani.ac.in}}
\affil[3]{School of Computer Science, KIIT University \\ \texttt{murari.mandalfcs@kiit.ac.in}}

\begin{document}

\maketitle

\begin{abstract}
Large language models (LLMs) are increasingly evaluated on mathematical reasoning, yet their robustness to equivalent problem representations remains poorly understood. In geometry, identical problems can be expressed in Euclidean, coordinate, or vector forms, but existing benchmarks report accuracy on fixed formats, implicitly assuming representation invariance and masking failures caused by representational changes alone. We propose \textbf{GeoRepEval}, a representation-aware evaluation framework that measures correctness, invariance, and consistency at the problem level across parallel formulations, combining strict answer matching, bootstrap confidence intervals, paired McNemar tests, representation-flip analyses, and regression controls for surface complexity. We prove that our Invariance@3 metric decomposes accuracy into robust and fragile components and is bounded by the weakest representation. Evaluating eleven LLMs on 158 curated high-school geometry problems (474 instances), we find accuracy gaps of up to 14 percentage points induced solely by representation choice. Vector formulations emerge as a consistent failure point, with Invariance@3 as low as 0.044 even after controlling for length and symbolic complexity. A convert-then-solve prompting intervention improves vector accuracy by up to 52 percentage points for high-capacity models, suggesting that failures reflect representation sensitivity rather than inability; however, low-capacity models show no gains, indicating deeper limitations. These results suggest that current models rely on representation-specific heuristics rather than abstract geometric reasoning. All datasets, prompts, and scripts are released at \url{https://github.com/vedjaw/GeoRepEval}.
\end{abstract}

\section{Introduction}
\label{sec:intro}

Large language models (LLMs) have demonstrated strong capabilities across reasoning-intensive tasks \cite{Brown2020, OpenAI2023}, including mathematical problem solving \cite{Cobbe2021, Hendrycks2021}, symbolic manipulation \cite{Lewkowycz2022}, and logical inference \cite{Wei2022}. Recent advances in scale, instruction tuning, and chain-of-thought prompting have led to substantial gains on benchmarks spanning arithmetic, algebra, and geometry \cite{Brown2020, Wei2022, Chowdhery2022, OpenAI2023}. However, growing evidence---sensitivity to prompt phrasing \cite{Zhu2023}, adversarial perturbations \cite{Goodfellow2015}, and surface-level rewording \cite{Ribeiro2020}---suggests LLM performance can be brittle under representational changes \cite{Srivastava2023}. Geometry provides a uniquely structured testbed: the same problem can be expressed via Euclidean, coordinate, or vector representations without altering its semantic content.

\paragraph{Research questions.}
This paper investigates whether modern LLMs exhibit representation-invariant geometric reasoning by addressing three questions:
\begin{enumerate}[nosep,label=\textbf{RQ\arabic*:}]
  \item Does performance vary systematically with representation choice, and can gaps be explained by surface features?
  \item Can explicit conversion-based prompting improve cross-representation consistency?
\end{enumerate}
We evaluate whether models produce correct and consistent answers when the same geometry problem is expressed in Euclidean, coordinate, and vector forms---representations that differ in symbolic density and linguistic structure but encode identical mathematical constraints \cite{Lewkowycz2022}.

Prior work focuses on aggregate accuracy over fixed-format benchmarks \cite{Cobbe2021, Hendrycks2021, Drori2022}, treating representation as incidental. Studies exploring paraphrasing robustness operate within single representational regimes \cite{Zhu2023}, and geometry benchmarks rarely provide parallel formulations of the same problem \cite{Trinh2024}. Current evaluations thus cannot determine whether LLMs internally normalize problems or rely on representation-specific heuristics.

We propose \textbf{GeoRepEval}, a representation-aware evaluation framework that constructs parallel Euclidean, coordinate, and vector versions of each problem and evaluates at the \emph{problem level}---comparing answers across all three variants. GeoRepEval introduces Invariance@3 (fraction of correct problems in all three representations) and Consistency@3 (fraction of problems with identical output), alongside McNemar tests, representation-flip analysis, and regression controls for surface features (token length, symbolic density).

We evaluate eleven LLMs spanning open-weight (7B--12B) and proprietary architectures on 474 curated geometry problems (158 instances per geometric representation). Euclidean--Vector accuracy gaps reach up to 14 pp (Claude-Haiku: 0.60 Euclid vs.\ 0.46 Vector), and Invariance@3---the fraction of problems solved correctly across all three representations---remains below 0.47 for all models except GPT-OSS-20B (0.665). At the low end, LLaMA-3.1-8B achieves only 0.044 Invariance@3, meaning fewer than 5\% of problems are solved correctly in all three forms. These gaps persist after regression controls for surface complexity ($p < 0.05$ for 8/11 models), and item-level analysis reveals vector formulations as the dominant single point of failure: the CCW pattern (correct on Euclidean and Coordinate, wrong on Vector) is 2--5$\times$ more frequent than WCC (Euclidean-only failure), indicating that models apply Euclidean reasoning ``templates'' fluently but produce brittle, error-prone vector procedures.

\paragraph{Contributions.}
\begin{itemize}[nosep]
    \item \textbf{GeoRepEval framework:} A statistically grounded, representation-aware evaluation framework with formally justified metrics (Properties~1--3) linking Invariance@3 and Consistency@3 to standard accuracy.
    \item \textbf{Parallel benchmark:} 158 core geometry problems each in Euclidean, coordinate, and vector form (474 instances per source) from standard textbooks, with verified semantic equivalence.
    \item \textbf{Empirical \& explanatory analysis:} Accuracy gaps up to 14 pp and Invariance@3 as low as 0.044; a qualitative analysis (\S\ref{sec:why_vector}) proposing procedural chain length, possible training distribution effects, and algebraic error compounding as plausible explanations for vector fragility.
    \item \textbf{Prompting intervention:} A convert-then-solve strategy that dramatically improves vector accuracy for mid/high-capacity models (up to 52 pp), suggesting failures are representation-specific; low-capacity models show no benefit, indicating deeper limitations.
\end{itemize}


\section{Related Work}
\label{sec:related}

\subsection{Mathematical Reasoning and Geometry with Neural Models}
Prior studies have evaluated LLM mathematical reasoning on benchmarks such as GSM8K and MATH \cite{Cobbe2021, Hendrycks2021, Drori2022, Lewkowycz2022, Polu2023}, with complementary surveys assessing model capabilities more broadly \cite{Qin2023, Huang2023}. Geometry-focused work has explored diagram understanding, symbolic reasoning, and coordinate-based formulations \cite{Seo2015, Trinh2024}. However, these evaluations typically assess each problem in a single canonical form, leaving it unclear whether success reflects abstract reasoning or sensitivity to representational encoding. GeoRepEval departs from this paradigm by evaluating the \emph{same} problem across Euclidean, coordinate, and vector formulations.

\subsection{Evaluation Frameworks for LLM Robustness}
Benchmarks such as BIG-bench \cite{Srivastava2023}, HELM \cite{Liang2022}, and Beyond-the-IID \cite{Chang2024} stress-test models across diverse conditions. Work on representation shift \cite{Zhang2024}, paraphrasing \cite{Ribeiro2020, Jiang2021}, and adversarial perturbations \cite{Min2023} has revealed sensitivity to superficial changes despite semantic equivalence. In mathematics, however, most benchmarks remain instance-level and lack aligned variants of the same problem, limiting invariance assessment. GeoRepEval fills this gap with problem-level, representation-aligned evaluation and paired statistical testing.

\subsection{Prompting and Reasoning Strategies}
Chain-of-thought prompting \cite{Wei2022}, self-consistency \cite{Wang2023}, tree-of-thought \cite{Yao2023}, and tool-augmented methods \cite{Schick2024} have improved reasoning performance, with zero-shot and least-to-most strategies \cite{Kojima2022, Zhou2023} unlocking latent ability without exemplars. However, these studies implicitly assume a fixed representation and focus on \emph{how} models reason rather than \emph{what form} reasoning is conditioned on. Our work isolates representation as a controlled variable, testing whether LLM reasoning is representation-invariant.


\section{Methodology}
\label{sec:method}

\begin{figure}[t]
  \centering
  \resizebox{0.95\columnwidth}{!}{%
  \begin{tikzpicture}[
      node distance=1.2cm and 0.8cm,
      box/.style={draw, rectangle, rounded corners, minimum width=2.4cm, minimum height=0.9cm, align=center, fill=blue!5, font=\small},
      arrow/.style={-{Stealth[scale=1.1]}, thick, draw=black!70},
      group/.style={draw=black!40, densely dashed, inner sep=12pt, rounded corners, fill=gray!3},
      grouplabel/.style={font=\bfseries\small, color=black!80, fill=white, inner sep=3pt}
  ]
  
      \node[box] (source) {1. Source\\Selection};
      \node[box, right=of source] (cat) {2. Category\\Balancing};
      \node[box, right=of cat] (gen) {3. Parallel\\Generation};
      \node[box, right=of gen] (filter) {4. Problem\\Filtering};
      
      \draw[arrow] (source) -- (cat);
      \draw[arrow] (cat) -- (gen);
      \draw[arrow] (gen) -- (filter);
      
      \begin{scope}[on background layer]
          \node[group, fit=(source)(cat)(gen)(filter)] (g1) {};
      \end{scope}
  
      \node[box, below=1.5cm of cat, xshift=-1.5cm, fill=orange!10] (euc) {Euclidean\\Geometry};
      \node[box, right=0.6cm of euc, fill=orange!10] (coord) {Coordinate\\Geometry};
      \node[box, right=0.6cm of coord, fill=orange!10] (vec) {Vector\\Geometry};
      
      \draw[arrow] (source.south) -- (euc.north);
      \draw[arrow] (cat.south) -- (coord.north);
      \draw[arrow] (filter.south) -- (vec.north);
  
      \begin{scope}[on background layer]
          \node[group, fit=(euc)(coord)(vec)] (g2) {};
      \end{scope}
  
      \node[box, below=1.2cm of euc, minimum width=3.8cm, fill=green!5] (scoring) {Normalization \&\\Strict Marking};
      \node[box, below=1.2cm of vec, minimum width=3.8cm, fill=green!5] (stats) {Invariance@3 \&\\Statistical Tests};
      
      \draw[arrow] (euc.south) -- (scoring.north);
      \draw[arrow] (coord.south) -- ($(scoring.north)!0.5!(stats.north)$);
      \draw[arrow] (vec.south) -- (stats.north);
      \draw[arrow] (scoring) -- (stats);
  
      \node[grouplabel, anchor=south west] at (g1.north west) {Dataset Construction};
      \node[grouplabel, anchor=south west] at (g2.north west) {Controlled Inference (LLMs)};
  
  \end{tikzpicture}
  }
  \caption{\textbf{GeoRepEval pipeline overview.} The framework natively constructs and tracks mathematically equivalent variants (Euclidean, Coordinate, and Vector) through parallel LLM inference to isolate true reasoning capacity from representation sensitivity.}
  \label{fig:system_diagram}
\end{figure}
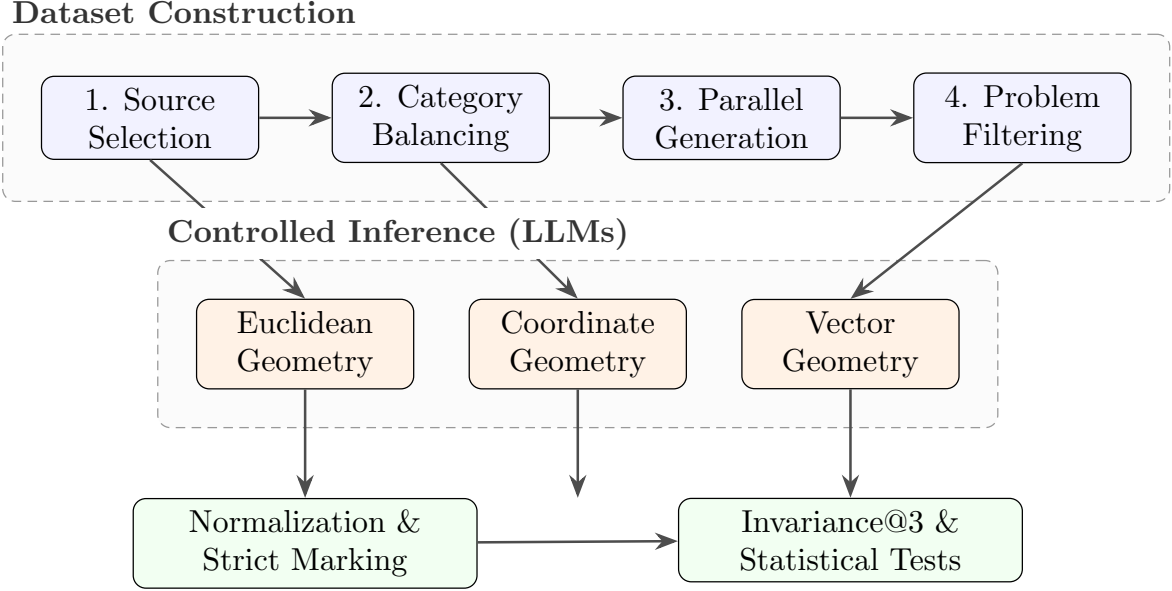

GeoRepEval probes whether LLMs exhibit invariant reasoning across equivalent geometric representations via four stages: dataset construction, controlled inference, structured scoring, and statistical analysis (Figure~\ref{fig:system_diagram}). Let $\mathcal{M}$ denote a language model. Each problem $p_i$ has gold answer $a_i^*$ and is expressed in representations $r \in \{\textsc{euc}, \textsc{coord}, \textsc{vec}\}$, yielding prompts $x_i^r$. The predicted answer $\hat{a}_i^r = \text{parse}(\mathcal{M}(x_i^r))$ is \emph{correct} if it matches $a_i^*$ after normalization (\S\ref{sec:method_comp3}). We capture this using a binary correctness indicator $c_i^r \in \{0, 1\}$ that flags whether the model successfully solved problem $p_i$ in its representation form $r$, defined formally as $c_i^r = \mathbf{1}[\text{normalize}(\hat{a}_i^r) = \text{normalize}(a_i^*)]$.

\subsection{Dataset Construction}
\label{sec:dataset_construction}

The construction pipeline proceeds in four stages:

\paragraph{Stage 1: Source Selection.}
Problems are drawn from three widely used Indian high-school and junior-college mathematics textbooks---NCERT (Classes IX--XII) \cite{NCERT2020}, RD~Sharma \cite{Sharma2019, Sharma2020}, and RS~Aggarwal \cite{Aggarwal2019, Aggarwal2020}---chosen for their emphasis on formal symbolic reasoning and well-defined geometric constraints.

\paragraph{Stage 2: Categorisation.}
Each problem is assigned to one of four categories: \emph{length/distance}, \emph{area/volume}, \emph{ratio/proportion}, and \emph{angle/direction}, with approximately balanced distribution.

\paragraph{Stage 3: Parallel Generation.}
Every core problem is instantiated into three equivalent forms: \textbf{Euclidean} (verbal/relational), \textbf{Coordinate} (Cartesian plane), and \textbf{Vector} (dot/cross-product operations). Each variant preserves identical constraints and solution paths. Variants are generated with Gemini Pro \cite{Gemini2023} and manually verified by a panel of mathematical experts.

\paragraph{Stage 4: Filtering.}
Problems with ambiguous phrasing, missing parameters, or non-exact answers are removed. After filtering, each source contributes 158 core problems (474 instances per source).

\subsection{Inference, Scoring, and Aggregation}
\label{sec:method_comp3}
\label{sec:method_comp2}

Each model is evaluated on all three representations under identical conditions, producing answers in structured JSON (Appendix~\ref{app:prompts}). GeoRepEval also includes a \emph{convert-then-solve} variant where models first rewrite coordinate/vector problems into Euclidean form. Predicted answers are compared against gold solutions using strict normalization (fractions, roots, $\pi$-expressions to canonical form):
\begin{equation}
  c_i^r = \mathbf{1}\bigl[\text{normalize}(\hat{a}_i^r) = \text{normalize}(a_i^*)\bigr].
\end{equation}
This avoids numeric tolerance issues while accepting equivalent forms (e.g., $\tfrac{3}{2}$ and $1.5$).

Results are aggregated across representations and problems. Beyond standard per-representation accuracy, GeoRepEval computes two higher-order metrics. \textbf{Invariance@3} measures the fraction of problems solved correctly in \emph{all three} representations:
\begin{equation}
  \text{Invariance@3} = \frac{1}{N}\sum_{i=1}^{N} c_i^{\textsc{euc}} \cdot c_i^{\textsc{coord}} \cdot c_i^{\textsc{vec}},
\end{equation}
where $N$ is the number of problems. \textbf{Consistency@3} measures the fraction of problems where the model produces the \emph{same} output (regardless of correctness) across all three representations:
\begin{equation}
  \text{Consistency@3} = \frac{1}{N}\sum_{i=1}^{N} \mathbf{1}\bigl[\hat{a}_i^{\textsc{euc}} = \hat{a}_i^{\textsc{coord}} = \hat{a}_i^{\textsc{vec}}\bigr].
\end{equation}
GeoRepEval also classifies each problem by its cross-representation correctness pattern (e.g., CCC, CCW, WWW), yielding eight categories for structured failure analysis.

\subsection{Formal Properties of Evaluation Metrics}
\label{sec:metric_properties}

We establish three structural properties that justify our two-metric design and clarify the relationship between accuracy, invariance, and consistency. Throughout, let $R = \{\textsc{euc}, \textsc{coord}, \textsc{vec}\}$ denote the set of representations, and recall that $c_i^r \in \{0,1\}$ is the correctness indicator for problem $i$ under representation $r$. A model is \emph{perfectly representation-invariant} if $\hat{a}_i^r = \hat{a}_i^{r'}$ for all problems $i$ and all representation pairs $(r, r')$.

\paragraph{Property 1 (Accuracy Decomposition).}
For any representation $r$, the per-representation accuracy decomposes as:
\begin{equation}
  \text{Acc}^r = \text{Invariance@3} + \underbrace{\frac{1}{N}\sum_{i=1}^{N} c_i^r \bigl(1 - \textstyle\prod_{r' \ne r} c_i^{r'}\bigr)}_{\text{fragile correctness } F^r}.
\end{equation}
\emph{Derivation.} By definition, $\text{Acc}^r = \frac{1}{N}\sum_i c_i^r$. We partition each $c_i^r$ according to whether the model also succeeds on every other representation: $c_i^r = c_i^r \prod_{r'\ne r} c_i^{r'} + c_i^r (1 - \prod_{r'\ne r} c_i^{r'})$. Summing over $i$ and dividing by $N$, the first term equals Invariance@3 and the second equals $F^r$. The fragile-correctness term $F^r$ captures problems solved in representation $r$ but not in all three, indicating reliance on representation-specific cues rather than robust reasoning.

\paragraph{Property 2 (Upper Bound).}
Invariance@3 is bounded above by the weakest per-representation accuracy:
\begin{equation}
  \text{Invariance@3} \leq \min_{r \in R}\; \text{Acc}^r.
\end{equation}
\emph{Derivation.} Because $c_i^r \in \{0,1\}$, the product $\prod_{r \in R} c_i^r \leq c_i^r$ for every $r$. Summing over problems, $\text{Invariance@3} = \frac{1}{N}\sum_i \prod_{r} c_i^r \leq \frac{1}{N}\sum_i c_i^r = \text{Acc}^r$. Since this holds for every $r$, it holds for the minimum. This formalises the \emph{weakest-link} effect: a model cannot achieve high invariance if it performs poorly on even one representation.

\paragraph{Property 3 (Consistency--Invariance Gap).}
Consistency@3 is always at least as large as Invariance@3:
\begin{equation}
  \text{Consistency@3} \geq \text{Invariance@3}.
\end{equation}
\emph{Derivation.} Let $S_i = \mathbf{1}[\hat{a}_i^{\textsc{euc}} = \hat{a}_i^{\textsc{coord}} = \hat{a}_i^{\textsc{vec}}]$ (same output) and $I_i = \prod_r c_i^r$ (all correct). If $I_i = 1$, then all three outputs match the gold answer, so they also match each other; hence $I_i \leq S_i$ for every $i$, and averaging gives the result. The gap $\text{Consistency@3} - \text{Invariance@3}$ measures the fraction of problems where the model produces the same \emph{wrong} answer across all three representations, indicating systematic errors orthogonal to representation sensitivity.

GeoRepEval prioritises paired, problem-level evaluation with semantically equivalent variants. These structural properties ensure the metrics are principled, and McNemar tests with bootstrap CIs (\S\ref{sec:results_qualitative}) confirm that observed differences are statistically systematic.


\section{Experimental Setup}
\label{sec:setup}

\subsection{Datasets}
\label{sec:datasets}

GeoRepEval was evaluated on a custom-curated geometry benchmark constructed from standard Indian high school and junior college mathematics textbooks, including NCERT \cite{NCERT2020}, RD~Sharma \cite{Sharma2019, Sharma2020}, and RS~Aggarwal \cite{Aggarwal2019, Aggarwal2020}. These sources emphasize formal symbolic reasoning and precise geometric constraints, making them well suited for robustness evaluation.

Each core problem was instantiated into \textbf{three mathematically equivalent representations}—Euclidean, coordinate, and vector geometry—while preserving identical constraints, quantities, and solution targets. Additional problems and parallel variants were generated and refined using the Gemini Pro model. All variants were then manually verified by a panel of mathematical experts to ensure correctness, semantic parallelism, and answer uniqueness. Problems with ambiguous phrasing or multiple valid interpretations were revised or removed. All retained problems admit exact numeric or symbolic answers.

\begin{table}[t]
  \centering
  \small
  \setlength{\tabcolsep}{4pt}
  \begin{tabularx}{\columnwidth}{l c X X}
    \hline
    \textbf{Source} & \textbf{\#Prob.} & \textbf{Categories} & \textbf{Repr.} \\
    \hline
    NCERT (IX--XII)       & 62 & Len., Area, Ratio, Angle & Euc., Coord., Vec. \\
    RD Sharma (X--XII)    & 53 & Len., Area, Ratio, Angle & Euc., Coord., Vec. \\
    RS Aggarwal           & 43 & Len., Area, Ratio, Angle & Euc., Coord., Vec. \\
    \hline
    \textbf{Total}        & \textbf{158} & & \\
    \hline
  \end{tabularx}
  \caption{\textbf{Dataset summary.} 158 core problems drawn from three textbook sources, each instantiated in three representations (474 total instances). All problems are at the high-school level.}
  \label{tab:datasets}
\end{table}

Using multiple textbook sources reduces source-specific bias, so that performance differences can be attributed to representation effects.

\subsection{Models and Evaluation Setup}
\label{sec:models}
\label{sec:metrics}
\label{sec:impl}

\begin{table}[t]
  \centering
  \small
  \setlength{\tabcolsep}{6pt}
  \begin{tabular}{l l c l l}
    \hline
    \textbf{Model} & \textbf{Architecture} & \textbf{Parameters} & \textbf{Provider} & \textbf{Access} \\
    \hline
    Claude Haiku 4.5 & Not Disclosed & Not Disclosed & Anthropic & Closed \\
    DeepSeek Chat & Decoder-only Transformer & Not Disclosed & DeepSeek & Open \\
    Gemini 2.5 Flash & Transformer-based architecture & Not Disclosed & Google & Closed \\
    Gemma 2 9B IT\textsuperscript{$\dagger$} & Decoder-only Transformer & 9B & Google & Open \\
    Gemma 3 12B IT & Decoder-only Transformer & 12B & Google & Open \\
    LLaMA 3.1 8B Instruct & Decoder-only Transformer & 8B & Meta & Open \\
    Mistral 7B Instruct & Decoder-only Transformer & 7B & Mistral & Open \\
    Qwen 2.5 7B Instruct & Decoder-only Transformer & 7B & Alibaba & Open \\
    GPT-5.1 & Decoder-only Transformer & Not Disclosed & OpenAI & Closed \\
    GPT-5.2 & Decoder-only Transformer & Not Disclosed & OpenAI & Closed \\
    GPT-OSS 20B & Decoder-only Transformer & 20B & OpenAI & Open \\
    \hline
  \end{tabular}
  \caption{\textbf{Large language models evaluated in this study.} All models were accessed via the OpenRouter API (\url{https://openrouter.ai}) in a zero-shot setting without any fine-tuning. \textsuperscript{$\dagger$}Gemma-2-9B produced valid responses for 110 of 158 problems due to output formatting failures; only valid responses are included in its evaluation.}
  \label{tab:model_summary}
\end{table}

Table~\ref{tab:model_summary} lists the evaluated models, spanning open-weight architectures (LLaMA \cite{Touvron2023}, Gemma, Qwen, Mistral) and proprietary systems (GPT \cite{OpenAI2024}, Gemini \cite{Gemini2023}, Claude).

\begin{table}[t]
  \centering
  \small
  \begin{tabular}{lll}
    \hline
    \textbf{Metric} & \textbf{Measures} & \textbf{Higher = Better?} \\
    \hline
    Accuracy (Euclid / Coord / Vec) & Per-representation accuracy & Yes \\
    Accuracy Gap & Best $-$ worst representation accuracy & No \\
    Invariance@3 & Correct across all three representations & Yes \\
    Consistency@3 & Identical output across all three & Yes \\
    Representation-Flip Categories & Problem-level patterns (CCC, CCW, etc.) & -- \\
    Pairwise Transfer / Coherence & Cross-representation success rates & Yes \\
    Bootstrap CI / McNemar's Test & Statistical reliability & -- \\
    Regression-Controlled Accuracy & Accuracy after surface-complexity controls & Yes \\
    Convert-then-Solve Accuracy & Accuracy under conversion prompting & Yes \\
    \hline
  \end{tabular}
  \caption{\textbf{Evaluation metrics.} See \S\ref{sec:method_comp3} and \S\ref{sec:metric_properties} for formal definitions and properties.}
  \label{tab:metrics}
\end{table}

All models were queried via the OpenRouter API using deterministic decoding (temperature~$= 0.0$, top-$p = 1.0$), zero-shot, with each prompt submitted once per problem--representation pair. Evaluation is based solely on the extracted numeric answer from structured JSON output (Appendix~\ref{app:prompts}). Non-conforming responses are treated as incorrect. Bootstrap resampling ($B = 10{,}000$) and McNemar tests use fixed seeds for reproducibility.

\paragraph{Manual verification of scoring.}
Although initial scoring used automated exact-match comparison, a systematic manual review was required to ensure fairness. Three recurring issues motivated this step: (i)~\emph{format mismatches}, where a model returned a mathematically equivalent but textually distinct answer (e.g., \texttt{"1/2"} vs.\ the stored gold \texttt{"0.5"}, or \texttt{"3.162"} vs.\ \texttt{"sqrt(10)"}); (ii)~\emph{unparseable symbolic forms}, such as \texttt{arccos(1/sqrt(3))} or \texttt{arccos(sqrt(2/3))}, which fell outside the coverage of our numeric-extraction regex and could not be compared programmatically; and (iii)~\emph{non-canonical expressions}, including unsimplified fractions (\texttt{"6/4"} vs.\ \texttt{"3/2"}), reordered terms (\texttt{"4*sqrt(3)"} vs.\ \texttt{"sqrt(48)"}), or extraneous text embedded in the JSON field. Each such case was adjudicated by a human annotator across all eleven models. In total, manual review covered approximately 15--20\% of the 5{,}214 evaluation instances (474 instances~$\times$~11 models), ensuring that reported metrics reflect genuine reasoning failures rather than answer-format artifacts.


\section{Results and Discussion}
\label{sec:results}

\subsection{Experiment 1 --- Accuracy by Representation}

\begin{figure*}[t]
  \centering
  \makebox[\textwidth][c]{\includegraphics[width=1.15\textwidth]{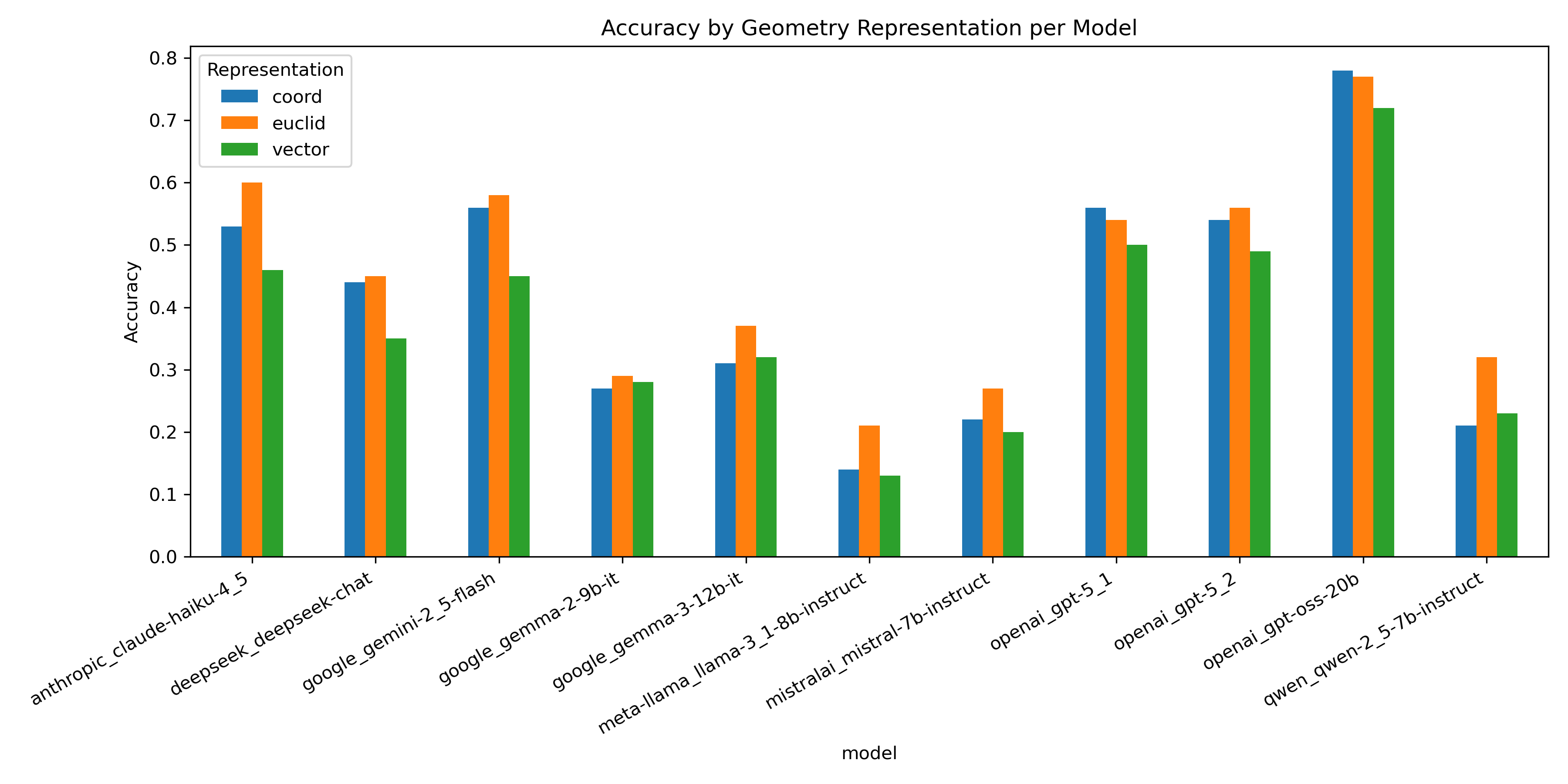}}
  \caption{\textbf{Accuracy by geometry representation across models.}
  Each group shows performance under Euclidean, Coordinate, and Vector formulations of the same problems.}
  \label{fig:accuracy_by_rep}
\end{figure*}

Figure~\ref{fig:accuracy_by_rep} shows that accuracy varies substantially across representations for all models. Euclidean formulations consistently yield the highest accuracy, vector formulations the lowest, with coordinate geometry between. These gaps persist across both open-weight and proprietary models, demonstrating that scale alone does not eliminate representation-induced degradation.

\subsection{Experiment 2 --- Invariance and Robustness}
\label{sec:results_main}

Table~\ref{tab:geometry_invariance_results} reports representation-wise accuracy, Invariance@3, and Consistency@3. Euclidean representations achieve the highest accuracy across nearly all models. Accuracy gaps reach up to 14 percentage points (e.g., Claude-Haiku: 0.60 Euclid vs.\ 0.46 Vector). Invariance@3 remains low for most models---LLaMA-3.1-8B achieves only 0.044---revealing that correct solutions across all three representations are rare. Models with higher overall accuracy tend to exhibit higher consistency, suggesting robustness and correctness are tightly coupled.

\begin{table}[t]
  \centering
  \footnotesize
  \setlength{\tabcolsep}{4pt}
  \begin{tabular}{l c c c l c c c}
    \hline
    \textbf{Model} &
    \textbf{Coord} &
    \textbf{Euclid} &
    \textbf{Vector} &
    \textbf{Best Rep.} &
    \textbf{Acc.\ Gap} &
    \textbf{Inv@3} &
    \textbf{Cons@3} \\
    \hline
    Claude Haiku 4.5 & 0.53 & 0.60 & 0.46 & Euclid & 0.14 & 0.401 & 0.631 \\
    DeepSeek Chat & 0.44 & 0.45 & 0.35 & Euclid & 0.10 & 0.266 & 0.424 \\
    Gemini 2.5 Flash & 0.56 & 0.58 & 0.45 & Euclid & 0.13 & 0.392 & 0.494 \\
    Gemma 2 9B IT & 0.27 & 0.29 & 0.28 & Euclid & 0.02 & 0.145 & 0.245 \\
    Gemma 3 12B IT & 0.31 & 0.37 & 0.32 & Euclid & 0.06 & 0.178 & 0.268 \\
    LLaMA 3.1 8B Inst. & 0.14 & 0.21 & 0.13 & Euclid & 0.08 & 0.044 & 0.101 \\
    Mistral 7B Inst. & 0.22 & 0.27 & 0.20 & Euclid & 0.07 & 0.089 & 0.197 \\
    GPT-5.1 & 0.56 & 0.54 & 0.50 & Coord & 0.06 & 0.468 & 0.791 \\
    GPT-5.2 & 0.54 & 0.56 & 0.49 & Euclid & 0.07 & 0.462 & 0.816 \\
    GPT-OSS 20B & 0.78 & 0.77 & 0.72 & Coord & 0.06 & 0.665 & 0.785 \\
    Qwen 2.5 7B Inst. & 0.21 & 0.32 & 0.23 & Euclid & 0.11 & 0.120 & 0.171 \\
    \hline
  \end{tabular}
  \caption{\textbf{Representation-wise accuracy, invariance, and consistency metrics across evaluated language models.} Accuracy Gap = best $-$ worst representation accuracy. Inv@3 = Invariance@3 (correct across all three). Cons@3 = Consistency@3 (identical output across all three). Inconsistency = $1 - \text{Cons@3}$.}
  \label{tab:geometry_invariance_results}
\end{table}


\subsection{Experiment 3 --- Representation-Flip Patterns}
\label{sec:results_second}

Item-level analysis (Figure~\ref{fig:rep_flip_patterns}, Appendix~\ref{app:extra_results}) reveals that WWW (all wrong) constitutes a large fraction, reflecting intrinsic difficulty. Among partial cases, CCW and CWC dominate---vector and coordinate are frequent single failure points, while WCC (Euclidean-only failure) is rare.


\subsection{Experiment 4 --- Pairwise Transfer}
\label{sec:results_analysis}

Vector is the dominant single-failure point across nearly all models, with failure counts 2--5$\times$ higher than Euclidean (Table~\ref{tab:pairwise_transfer}, Appendix~\ref{app:extra_results}). Transfer is strongest for Euclidean--Coordinate when Vector fails (EC$|$V up to 0.23), while CV$|$E and EV$|$C remain below 0.06. EC coherence consistently exceeds CV coherence, indicating more aligned reasoning pathways between Euclidean and Coordinate representations.


\subsection{Experiment 5 --- Statistical Reliability}
\label{sec:results_qualitative}

Bootstrap confidence intervals confirm that representation-induced gaps are statistically stable for mid- and high-capacity models, with non-overlapping intervals for Euclidean--Vector comparisons (Claude-Haiku, Gemini-2.5, GPT-5.2, GPT-OSS-20B). McNemar's paired tests \cite{McNemar1947} confirm significant effects for Euclidean--Vector and Coordinate--Vector comparisons ($p < 0.05$), while Euclidean--Coordinate differences are less consistently significant, reinforcing Vector as the primary divergence source.


\subsection{Qualitative Failure Analysis}
\label{sec:qualitative}

We examine representative failure cases. A triangle-area problem solved correctly in Euclidean form ($\frac{1}{2} \times 5 \times 12 = 30$) frequently produces algebraic errors in vector form ($\frac{1}{2}|\vec{AB} \times \vec{AC}|$), with models misidentifying vectors or miscomputing cross products. Similarly, distance problems solved via the coordinate formula fail when reformulated as $\|\vec{v}\|$, with models misapplying norms. These patterns suggest stronger associations with Euclidean/coordinate ``templates'' from training, while vector reasoning relies on fragile procedural chains.

\subsection{Analysis: Why Vector Representations Are Harder}
\label{sec:why_vector}

We propose four plausible explanations for the consistent underperformance on vector formulations. These are offered as conjectures supported by preliminary evidence from our experiments, not as confirmatory claims.

\paragraph{H1: Surface Complexity.}
Vector formulations are inherently longer and more symbol-heavy (e.g., requiring explicit $\hat{i}, \hat{j}, \hat{k}$ notation), and sequence length may correlate with generation difficulty. Our logistic regression analysis (\S\ref{sec:results_qualitative}) includes controls for prompt token length and symbolic density; the representation indicator variable remains significant ($p < 0.05$ for 8 of 11 models) even with these covariates, providing preliminary evidence that the gap is not fully attributable to surface features alone. However, we cannot rule out other confounds correlated with representation choice, so surface complexity remains a plausible partial contributor rather than a confirmed explanation.

\paragraph{H2: Procedural Chain Length.}
Vector solutions typically demand longer, multi-step algebraic chains. For example, computing the area of a right triangle in Euclidean form requires one operation ($0.5 \times b \times h$), whereas the vector form requires computing two displacement vectors (6 subtractions), evaluating their cross product (6 multiplications, 3 subtractions), and calculating the magnitude (3 squares, 1 square root). This extended procedural chain plausibly increases the probability of a spontaneous calculation error before reaching the final answer, though we do not have a direct experimental control isolating chain length from other factors.

\paragraph{H3: Training Data Distribution (Speculative).}
Since training corpora for the evaluated models are not publicly documented, we cannot make direct claims about their composition. However, it is plausible that Euclidean and coordinate geometry---standard topics in secondary-school curricula---are more heavily represented in typical web and textbook data than vector geometry, which is generally introduced at the advanced high-school or college level. The success of the convert-then-solve intervention (\S\ref{sec:convert_then_solve})---where vector accuracy improves by up to 52~pp simply by prompting the model to translate to Euclidean first---is consistent with this hypothesis, suggesting that models may possess the underlying conceptual understanding but lack well-practiced procedural routines for vector arithmetic. We stress that this remains speculative in the absence of access to training data.

\paragraph{H4: Compounding Algebraic Errors.}
Unlike Euclidean geometry, which often relies on proportional reasoning, vector geometry requires strict adherence to algebraic rules (signs, component-wise arithmetic). Qualitative inspection reveals that models frequently state the correct high-level vector formulas (e.g., $\vec{u} \cdot \vec{v} = 0$ for orthogonality) but make sign errors when expanding the components. The dominance of the CCW pattern (Correct Euclidean and Coordinate, Wrong Vector) in the partial failure modes (\S\ref{sec:results_second}) is consistent with this explanation, suggesting that the models may possess adequate conceptual understanding while their symbolic execution of vector mechanics remains brittle. However, we note this is observational evidence and a more controlled study isolating algebraic error propagation would be needed to confirm this hypothesis.

\subsection{Convert-Then-Solve: A Prompting Intervention}
\label{sec:convert_then_solve}

RQ3 asks whether conversion-based prompting improves invariance. We evaluate CTS on a representative subset of six models selected to span the full performance spectrum---from low-capacity (LLaMA-3.1-8B) through mid-range (Qwen-2.5-7B, DeepSeek-Chat) to high-capacity (Claude-Haiku, Gemini-2.5-Flash, GPT-OSS-20B)---as the CTS protocol requires two-stage inference (conversion + solving), effectively doubling per-instance API cost. CTS dramatically improves accuracy for mid- and high-capacity models---vector accuracy jumps by up to 52 pp (Gemini-2.5-Flash: 0.45$\to$0.97), and accuracy gaps narrow from 14 pp to 2--3 pp---confirming that direct-evaluation failures reflect representation sensitivity rather than inability to solve the underlying problems. Crucially, LLaMA-3.1-8B shows no meaningful gains (vector: 0.13$\to$0.16), indicating that conversion scaffolding cannot compensate for fundamental capacity limitations. Full results appear in Appendix~\ref{app:cts_results}.

\subsection{Discussion}
\label{sec:discussion}

\textbf{Synthesis.}
The experiments provide evidence that LLM performance varies significantly across equivalent geometric representations (RQ2), with the Consistency--Invariance gap (Property~3) revealing representation-specific rather than systematic errors (RQ1). Convert-then-solve dramatically improves accuracy for capable models but fails for low-capacity ones (RQ3), suggesting that representation sensitivity is the primary bottleneck for strong models, while weaker models face deeper reasoning limitations.

\textbf{Implications.}
High accuracy on a single representation does not imply abstract reasoning competence. Current models show stronger inductive biases toward forms prevalent in training data and requiring shorter procedural chains, motivating representation-aware evaluation alongside standard benchmarks. For safety-critical applications, representation choice materially affects reliability; GeoRepEval enables systematic risk diagnosis.

\section{Conclusion}
\label{sec:conclusion}

We introduced GeoRepEval, a statistically grounded evaluation framework with formally justified metrics (Properties~1--3) for measuring representation invariance in LLM geometric reasoning. Across eleven models on 158 curated problems (474 instances per source), we find performance gaps of up to 14 pp induced solely by representation choice, with Invariance@3 below 0.47 for most models. A qualitative analysis proposes procedural chain length, possible training distribution effects, and compounding algebraic errors as plausible explanations for vector fragility. A convert-then-solve intervention dramatically improves vector accuracy for capable models (up to 52 pp), suggesting that failures are representation-specific rather than capability-based, while low-capacity models show no gains---revealing a capacity-dependent boundary for prompting interventions.

These results suggest that current LLMs may rely partly on representation-specific heuristics rather than fully abstract geometric reasoning, revealing fragility invisible to single-format benchmarks. Future work should extend GeoRepEval to chain-of-thought and few-shot regimes, additional mathematical domains, and training-time interventions for representation robustness.

\section*{Acknowledgments}

We thank our institute for GPU infrastructure, and OpenRouter for unified API access across all models. We further thank Shubh Srivastava and Aayan Ansari for their contribution to the human evaluation of a total of 4500 problems. We also thank peers for discussions that refined the experimental design.

\noindent\textbf{AI Use Statement.}
The authors are responsible for all content. AI tools were used for grammar checking and generating initial problem variants (manually verified). All scientific content is the authors' own.

\section*{Limitations}
\label{sec:limitations}

\begin{itemize}[nosep]
    \item \textbf{Prompting scope:} We use zero-shot structured prompting; CoT, few-shot, and self-consistency strategies may partially mitigate representation sensitivity.
    \item \textbf{Dataset:} 158 core problems from Indian high-school curricula; broader coverage of curricula, difficulty levels, and problem types would strengthen generalizability.
    \item \textbf{Methodology:} Exact-answer matching may undercount partially correct reasoning. Subtle phrasing differences across representations could influence results.
    \item \textbf{Benchmark construction:} Parallel problem variants were generated with Gemini Pro and manually verified; since Gemini 2.5 Flash is among the evaluated models, there is a potential contamination risk from within the same model family. However, the generation model (Gemini Pro) and evaluation model (Gemini 2.5 Flash) are distinct, and all generated variants were independently verified for correctness and parallelism.
    \item \textbf{Scope of claims:} Our findings suggest reliance on representation-specific heuristics but do not prove absence of abstract reasoning. Results are specific to text-only geometry and may not generalize to multimodal, multilingual, or other mathematical domains.
\end{itemize}

\section*{Ethics Statement}
This work evaluates publicly available language models on geometry problems from published textbooks. No human subjects were involved, and no personal data was collected. The evaluation framework and all datasets are available at \url{https://github.com/vedjaw/GeoRepEval} to support reproducibility.

\appendix
\section*{Appendix}

\section{Full Prompt Templates}
\label{app:prompts}

\begin{figure}[h]
\small
\begin{tcolorbox}[
  colback=gray!5,
  colframe=black!75,
  title={\textbf{Prompt Template}},
  fonttitle=\small\bfseries,
  boxrule=0.5pt,
  arc=2pt,
  left=4pt, right=4pt, top=2pt, bottom=2pt
]
\begin{Verbatim}[breaklines,breakanywhere,fontsize=\scriptsize]
You are solving a geometry problem.
You MUST follow ALL rules exactly.

---------- RULES ----------
1. Output MUST be valid JSON only.
2. Do NOT include markdown formatting,
   code blocks, or extra text.
3. The JSON must contain EXACTLY two keys:
   - "reasoning"
   - "numeric_answer"
4. "reasoning":
   - Must clearly explain the math steps.
   - May use words, symbols, equations.
   - MUST NOT contain literal line breaks.
     Use the escaped string "\\n".
5. "numeric_answer":
   - MUST be a string.
   - MUST contain ONLY the final numeric
     answer. No words or units.
6. Allowed formats for "numeric_answer":
   "5", "3/2", "sqrt(8)", "2*sqrt(5)",
   "8*pi"
7. If any rule is violated, the output
   is considered WRONG.

-------- OUTPUT FORMAT --------
{
  "reasoning": "<reasoning using \\n>",
  "numeric_answer": "<final answer>"
}

---------- PROBLEM ----------
{problem}
\end{Verbatim}
\end{tcolorbox}
\end{figure}

\section{Additional Experimental Results}
\label{app:extra_results}

\begin{figure}[h]
  \centering
  \includegraphics[width=\textwidth]{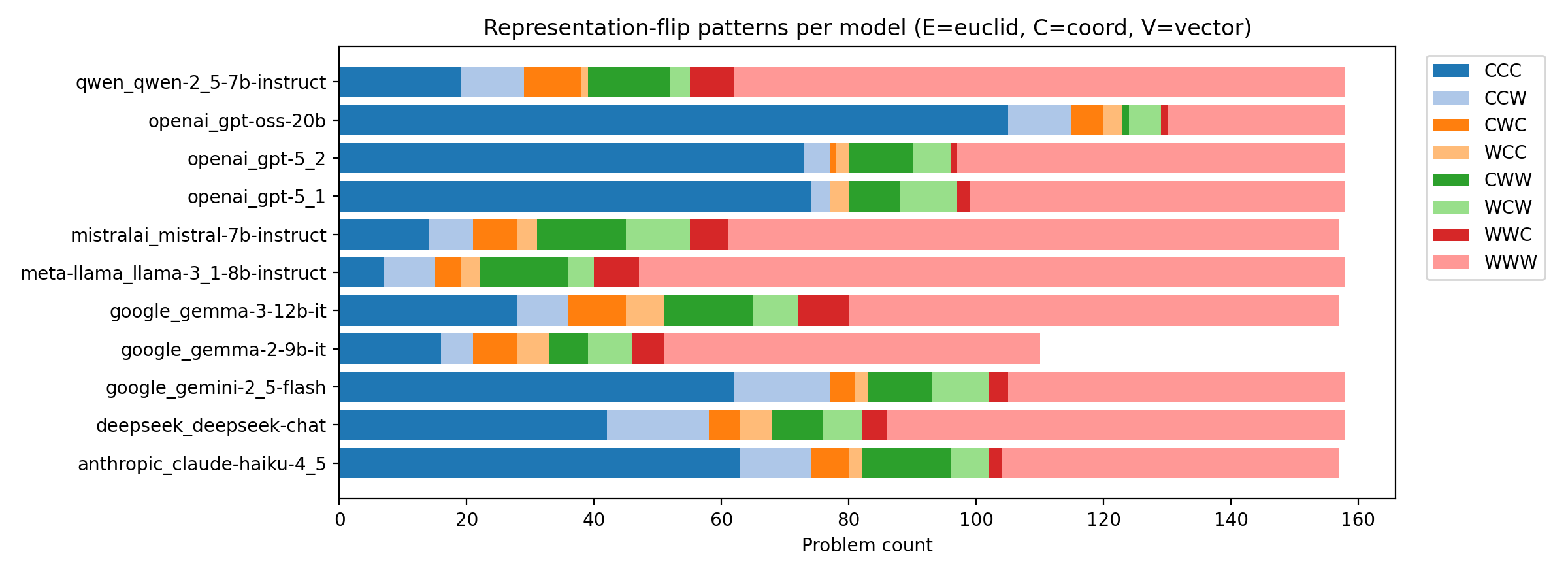}
  \caption{\textbf{Representation-flip patterns.} Stacked bars show problem-level correctness patterns (E, C, V) for each model.}
  \label{fig:rep_flip_patterns}
\end{figure}

\begin{table}[h]
  \centering
  \small
  \setlength{\tabcolsep}{4pt}
  \begin{tabular}{l c c c c c c c c c}
    \hline
    \textbf{Model} &
    \textbf{\#Prob} &
    \textbf{Fail-E} &
    \textbf{Fail-C} &
    \textbf{Fail-V} &
    \textbf{EC$|$V} &
    \textbf{CV$|$E} &
    \textbf{EV$|$C} &
    \textbf{EC-Co} &
    \textbf{CV-Co} \\
    \hline
    Claude-Haiku-4.5 & 157 & 2 & 6 & 11 & 0.13 & 0.03 & 0.08 & 0.82 & 0.84 \\
    DeepSeek-Chat & 158 & 5 & 5 & 16 & 0.16 & 0.06 & 0.06 & 0.85 & 0.80 \\
    Gemini-2.5-Flash & 158 & 2 & 4 & 15 & 0.17 & 0.03 & 0.06 & 0.84 & 0.80 \\
    Gemma-2-9B & 110 & 5 & 7 & 5 & 0.06 & 0.07 & 0.09 & 0.77 & 0.78 \\
    Gemma-3-12B & 157 & 6 & 9 & 8 & 0.08 & 0.06 & 0.08 & 0.77 & 0.80 \\
    LLaMA-3.1-8B & 158 & 3 & 4 & 8 & 0.06 & 0.02 & 0.03 & 0.84 & 0.85 \\
    Mistral-7B & 157 & 3 & 7 & 7 & 0.06 & 0.03 & 0.06 & 0.78 & 0.81 \\
    GPT-5.1 & 158 & 3 & 0 & 3 & 0.04 & 0.04 & 0.00 & 0.87 & 0.91 \\
    GPT-5.2 & 158 & 2 & 1 & 4 & 0.05 & 0.03 & 0.01 & 0.88 & 0.92 \\
    GPT-OSS-20B & 158 & 3 & 5 & 10 & 0.23 & 0.08 & 0.14 & 0.91 & 0.87 \\
    Qwen-2.5-7B & 158 & 1 & 9 & 10 & 0.08 & 0.01 & 0.07 & 0.84 & 0.82 \\
    \hline
  \end{tabular}
  \caption{\textbf{Pairwise transfer and coherence analysis.} Transfer rates (EC$|$V, CV$|$E, EV$|$C) and pairwise coherence (EC-Co, CV-Co).}
  \label{tab:pairwise_transfer}
\end{table}

\begin{figure}[h]
  \includegraphics[width=\columnwidth]{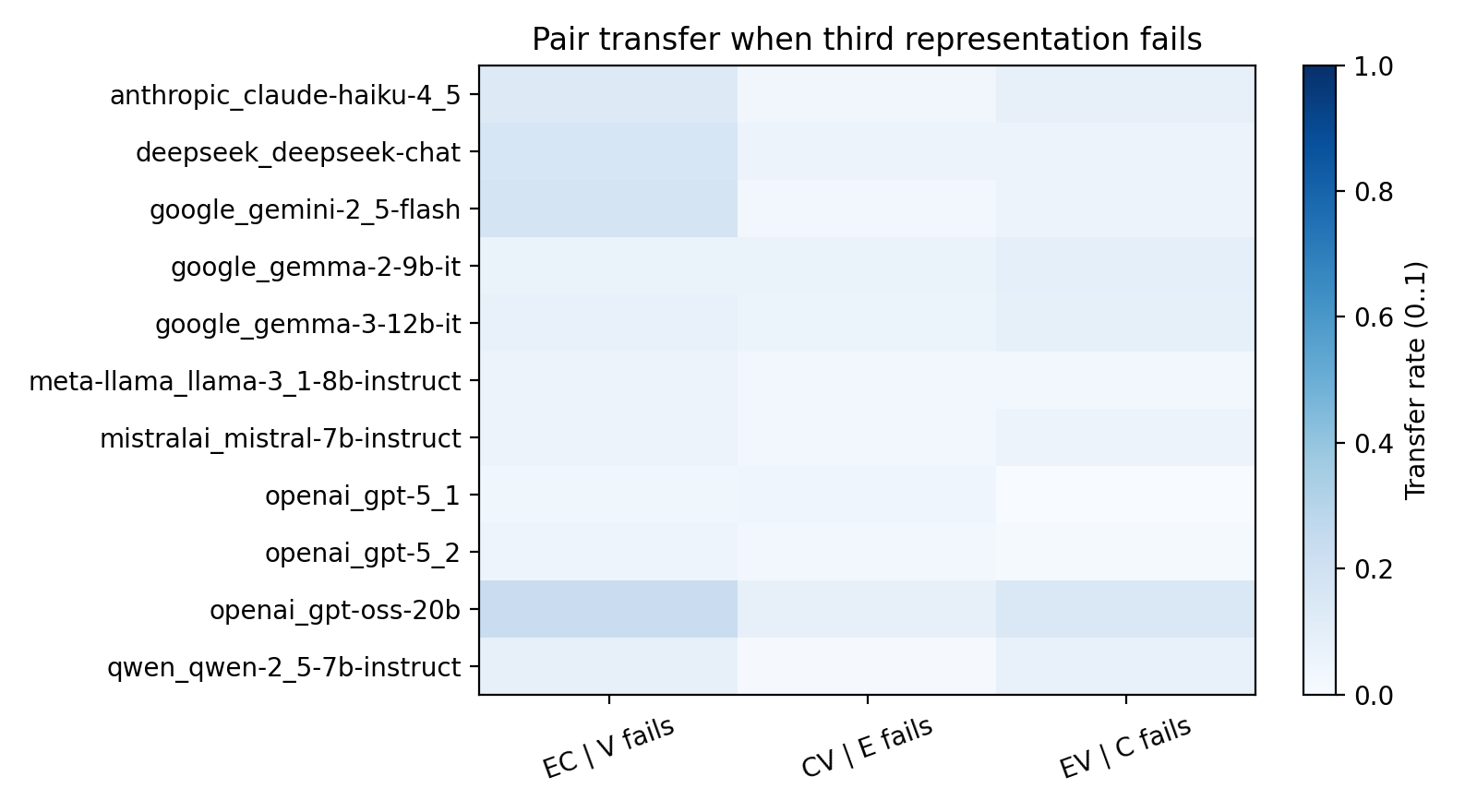}
  \caption{\textbf{Pairwise transfer rates.} Heatmap of how often two representations succeed when the third fails.}
  \label{fig:breakdown}
\end{figure}

Table~\ref{tab:rep_flip_counts} reports problem-level correctness pattern counts (CCC--WWW) across Euclidean, Coordinate, and Vector representations for all evaluated models.

\begin{table}[t]
  \centering
  \small
  \setlength{\tabcolsep}{4pt}
  \begin{tabular}{l c c c c c c c c c}
    \hline
    \textbf{Model} &
    \textbf{\#Prob} &
    \textbf{CCC} &
    \textbf{CCW} &
    \textbf{CWC} &
    \textbf{WCC} &
    \textbf{CWW} &
    \textbf{WCW} &
    \textbf{WWC} &
    \textbf{WWW} \\
    \hline
    Claude-Haiku-4.5 & 157 & 63 & 11 & 6 & 2 & 14 & 6 & 2 & 53 \\
    DeepSeek-Chat & 158 & 42 & 16 & 5 & 5 & 8 & 6 & 4 & 72 \\
    Gemini-2.5-Flash & 158 & 62 & 15 & 4 & 2 & 10 & 9 & 3 & 53 \\
    Gemma-2-9B & 110 & 16 & 5 & 7 & 5 & 6 & 7 & 5 & 59 \\
    Gemma-3-12B & 157 & 28 & 8 & 9 & 6 & 14 & 7 & 8 & 77 \\
    LLaMA-3.1-8B & 158 & 7 & 8 & 4 & 3 & 14 & 4 & 7 & 111 \\
    Mistral-7B & 157 & 14 & 7 & 7 & 3 & 14 & 10 & 6 & 96 \\
    GPT-5.1 & 158 & 74 & 3 & 0 & 3 & 8 & 9 & 2 & 59 \\
    GPT-5.2 & 158 & 73 & 4 & 1 & 2 & 10 & 6 & 1 & 61 \\
    GPT-OSS-20B & 158 & 105 & 10 & 5 & 3 & 1 & 5 & 1 & 28 \\
    Qwen-2.5-7B & 158 & 19 & 10 & 9 & 1 & 13 & 3 & 7 & 96 \\
    \hline
  \end{tabular}
  \caption{\textbf{Problem-level representation-flip pattern counts.}
  Each triplet (E,C,V) records correctness (C/W) across Euclidean, Coordinate, and Vector formulations.}
  \label{tab:rep_flip_counts}
\end{table}

\begin{figure}[t]
  \centering
  \includegraphics[width=\columnwidth]{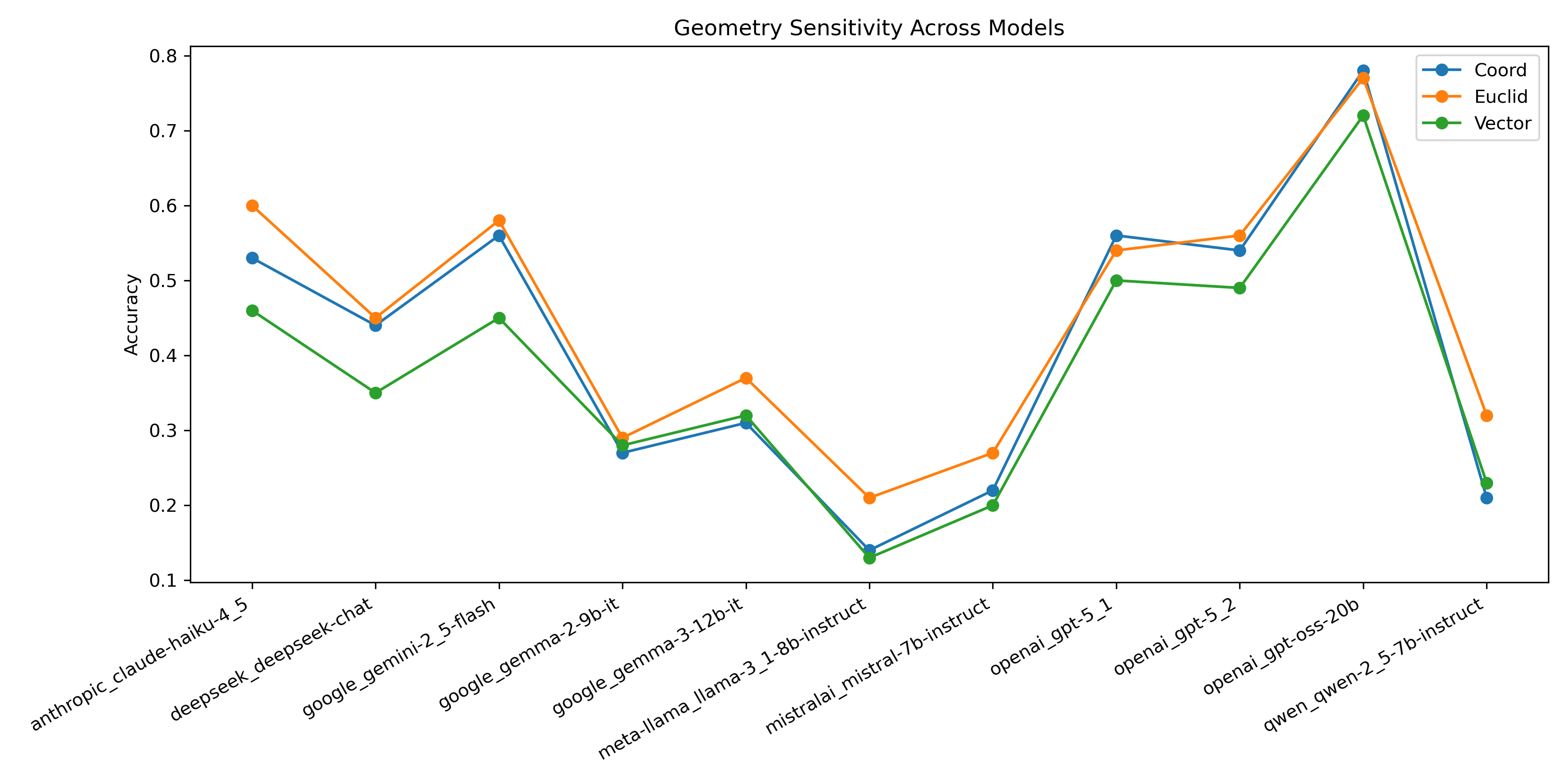}
  \caption{\textbf{Geometry sensitivity across models.}
  Representation-wise accuracy variation for each evaluated model under Euclidean, Coordinate, and Vector formulations, illustrating differential robustness to problem representation.}
  \label{fig:geometry_sensitivity}
\end{figure}

Figure~\ref{fig:geometry_sensitivity} visualizes representation-wise accuracy trends across models, highlighting systematic sensitivity to Euclidean, coordinate, and vector formulations.

\subsection{Convert-Then-Solve Results}
\label{app:cts_results}

Table~\ref{tab:cts_results} reports accuracy under the convert-then-solve (CTS) prompting intervention, where models first rewrite coordinate and vector problems into Euclidean form before solving. We evaluate CTS on a \emph{strategically selected} subset of six models chosen to maximize coverage across the performance distribution observed in the direct evaluation (Table~\ref{tab:geometry_invariance_results}): (1)~two high-capacity models with the strongest direct-evaluation accuracy (Claude-Haiku-4.5, Gemini-2.5-Flash), (2)~two mid-capacity models representing the median range (DeepSeek-Chat, GPT-OSS-20B), and (3)~two lower-capacity models spanning the weakest tiers (LLaMA-3.1-8B, Qwen-2.5-7B). This stratified selection ensures that the CTS analysis captures behavior across the full capability spectrum while managing the computational overhead: the CTS protocol requires two sequential inference passes per problem (conversion followed by solving), effectively doubling per-instance API cost. Running the full eleven-model suite under CTS would have tripled the total API budget without proportional information gain, as the six selected models already span the relevant performance and architectural diversity.

Compared to the direct evaluation, CTS substantially narrows accuracy gaps for high-capacity models: Claude-Haiku achieves near-ceiling accuracy across all representations (0.94--0.96), and Gemini-2.5-Flash reaches 0.95--0.97. The accuracy gap drops from 14 pp to 2 pp for these models. However, LLaMA-3.1-8B shows no meaningful improvement under CTS (Invariance@3 = 0.026), confirming that representational scaffolding benefits primarily models with sufficient baseline reasoning capability---a finding consistent with prior work showing that prompting interventions do not compensate for fundamental capacity limitations \cite{Wei2022}.

\begin{table}[h]
  \centering
  \small
  \setlength{\tabcolsep}{5pt}
  \begin{tabular}{l c c c c c c}
    \hline
    \textbf{Model} &
    \textbf{Coord} &
    \textbf{Euclid} &
    \textbf{Vector} &
    \textbf{Acc.~Gap} &
    \textbf{Inv@3} &
    \textbf{Cons@3} \\
    \hline
    Claude-Haiku-4.5 & 0.96 & 0.94 & 0.96 & 0.02 & 0.918 & 0.741 \\
    DeepSeek-Chat & 0.90 & 0.91 & 0.88 & 0.03 & 0.791 & 0.646 \\
    Gemini-2.5-Flash & 0.96 & 0.95 & 0.97 & 0.02 & 0.948 & 0.812 \\
    LLaMA-3.1-8B & 0.13 & 0.23 & 0.16 & 0.10 & 0.026 & 0.082 \\
    GPT-OSS-20B & 0.89 & 0.87 & 0.86 & 0.03 & 0.772 & 0.519 \\
    Qwen-2.5-7B & 0.83 & 0.77 & 0.78 & 0.06 & 0.596 & 0.443 \\
    \hline
  \end{tabular}
  \caption{\textbf{Convert-then-solve (CTS) accuracy.} Models first rewrite coordinate/vector problems into Euclidean form, then solve. Compared to direct evaluation (Table~\ref{tab:geometry_invariance_results}), CTS substantially narrows representation gaps for high-capacity models while having minimal effect on low-capacity ones.}
  \label{tab:cts_results}
\end{table}

\begin{figure}[h]
  \centering
  \includegraphics[width=\textwidth]{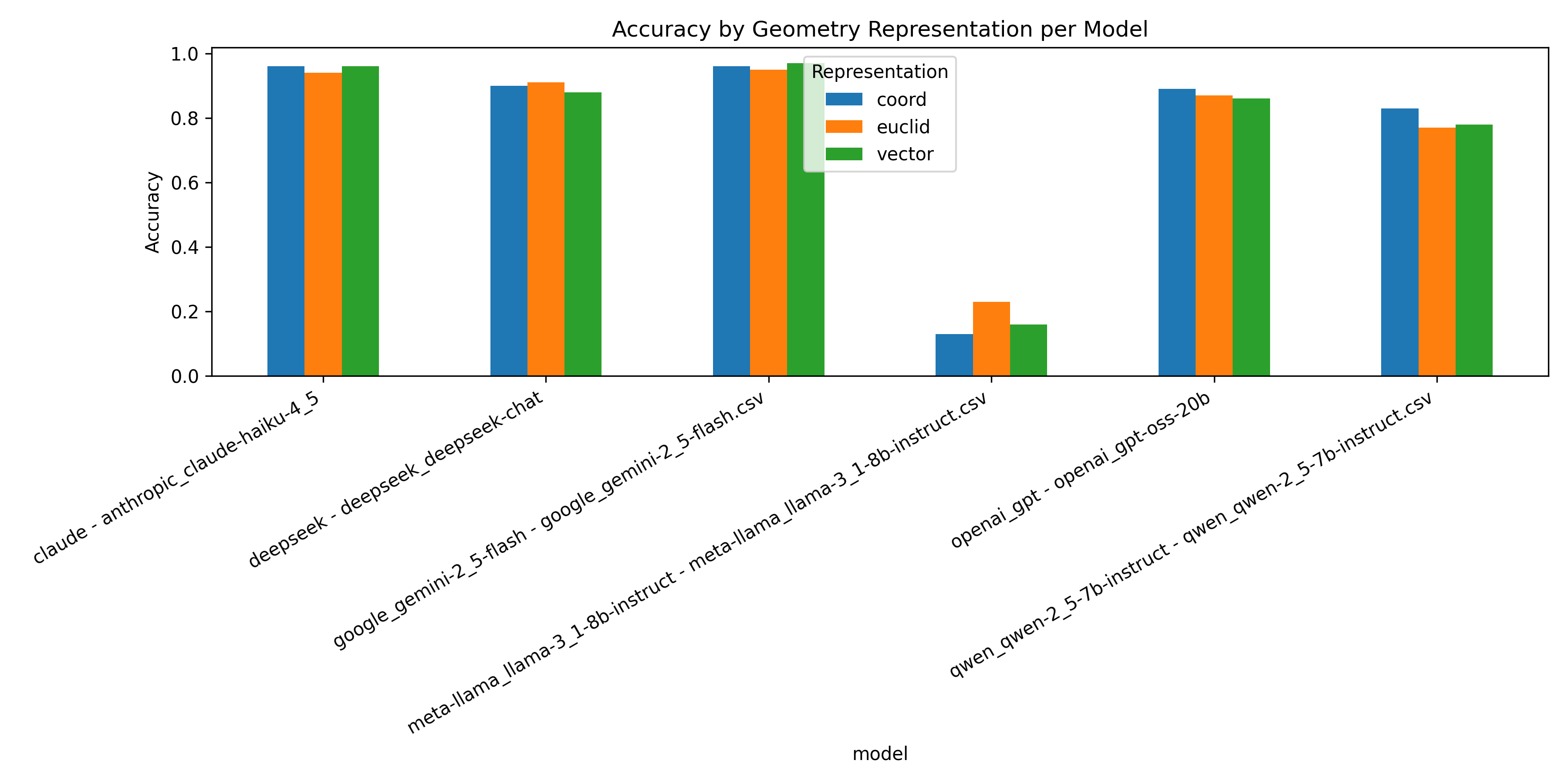}
  \caption{\textbf{Convert-then-solve accuracy by representation.} Accuracy under the CTS intervention for each model across Euclidean, Coordinate, and Vector formulations. Compared to direct evaluation (Figure~\ref{fig:accuracy_by_rep}), representation gaps are substantially reduced for high-capacity models.}
  \label{fig:cts_accuracy}
\end{figure}

\begin{figure}[h]
  \centering
  \includegraphics[width=\textwidth]{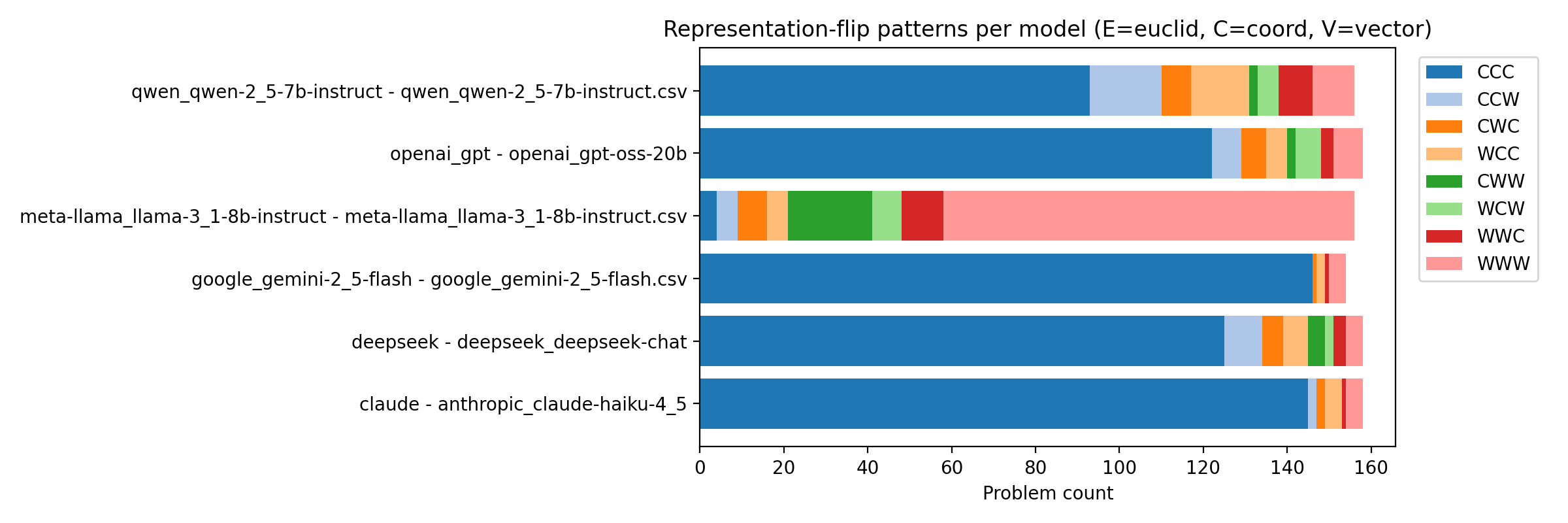}
  \caption{\textbf{Representation-flip patterns under convert-then-solve.} Stacked bars show problem-level correctness patterns (E, C, V) for each model after CTS prompting. Compared to direct evaluation (Figure~\ref{fig:rep_flip_patterns}), high-capacity models show a marked shift from partial-correctness patterns (CCW, CWC) toward fully invariant solutions (CCC).}
  \label{fig:cts_rep_flip}
\end{figure}

Figures~\ref{fig:cts_accuracy} and~\ref{fig:cts_rep_flip} visualize the CTS results. The bar chart (Figure~\ref{fig:cts_accuracy}) shows that representation gaps are dramatically reduced for most models under CTS, with Claude-Haiku, Gemini-2.5-Flash, and DeepSeek approaching near-uniform accuracy across all three representations. The representation-flip analysis (Figure~\ref{fig:cts_rep_flip}) reveals the underlying mechanism: CTS converts a large proportion of CCW and CWC patterns (single-representation failures) into CCC (all correct), substantially increasing the fraction of fully invariant solutions for high-capacity models. For LLaMA-3.1-8B, the WWW category remains dominant, confirming that conversion scaffolding cannot overcome fundamental reasoning limitations.

\section{Statistical Significance Tests}
\label{app:stat_tests}

This appendix reports the statistical tests used to assess whether observed
performance differences across geometry representations are reliable and
not attributable to sampling noise. Following common practice for paired
classification settings, we employ McNemar’s test to compare model correctness
between representation pairs at the problem level. Statistical significance
is evaluated at a threshold of $p < 0.05$.

\subsection{Paired McNemar Significance Tests}

Table~\ref{tab:mcnemar_tests} reports McNemar $p$-values and corresponding
$\chi^2$ statistics for all pairwise comparisons between Euclidean (E),
Coordinate (C), and Vector (V) representations. Significant results indicate
systematic representation-dependent differences in model behavior.

\begin{table}[t]
  \centering
  \small
  \setlength{\tabcolsep}{4pt}
  \begin{tabular}{l c c c c c c}
    \hline
    \textbf{Model} &
    \textbf{$p$(C--E)} & $\chi^2$ &
    \textbf{$p$(C--V)} & $\chi^2$ &
    \textbf{$p$(E--V)} & $\chi^2$ \\
    \hline
    Claude-Haiku-4.5 & 0.036 & 4.32 & 0.108 & 2.56 & \textbf{0.0001} & 13.79 \\
    DeepSeek-Chat & 0.839 & 0.04 & \textbf{0.029} & 4.65 & \textbf{0.014} & 5.94 \\
    Gemini-2.5-Flash & 0.690 & 0.16 & \textbf{0.003} & 8.26 & \textbf{0.0003} & 12.03 \\
    Gemma-2-9B & 1.000 & 0.00 & 1.000 & 0.00 & 1.000 & 0.00 \\
    Gemma-3-12B & 0.132 & 2.25 & 0.860 & 0.03 & 0.243 & 1.36 \\
    LLaMA-3.1-8B & \textbf{0.043} & 4.00 & 1.000 & 0.00 & \textbf{0.050} & 3.78 \\
    Mistral-7B & 0.229 & 1.44 & 0.585 & 0.30 & \textbf{0.043} & 4.03 \\
    GPT-5.1 & 0.503 & 0.45 & \textbf{0.013} & 5.79 & 0.210 & 1.56 \\
    GPT-5.2 & 0.648 & 0.21 & \textbf{0.039} & 4.08 & \textbf{0.013} & 5.88 \\
    GPT-OSS-20B & 0.791 & 0.07 & 0.078 & 3.05 & 0.118 & 2.40 \\
    Qwen-2.5-7B & \textbf{0.001} & 11.12 & 0.711 & 0.14 & \textbf{0.011} & 6.32 \\
    \hline
  \end{tabular}
  \caption{\textbf{McNemar paired significance tests across geometry representations.}
  Bold $p$-values indicate statistically significant differences ($p < 0.05$).}
  \label{tab:mcnemar_tests}
\end{table}

\subsection{Disagreement Structure Analysis}

To contextualize the significance results, Table~\ref{tab:mcnemar_counts}
reports the paired disagreement counts underlying McNemar’s test.
Here, $b$ denotes the number of problems solved correctly by the first
representation but not the second, while $c$ denotes the reverse.
Asymmetric $b$/$c$ counts explain the direction and magnitude of observed
significance.

\begin{table}[t]
  \centering
  \small
  \setlength{\tabcolsep}{4pt}
  \begin{tabular}{l c c c c c c}
    \hline
    \textbf{Model} &
    \textbf{$b_{CE}$} & $c_{CE}$ &
    \textbf{$b_{CV}$} & $c_{CV}$ &
    \textbf{$b_{EV}$} & $c_{EV}$ \\
    \hline
    Claude-Haiku-4.5 & 8 & 20 & 17 & 8 & 25 & 4 \\
    DeepSeek-Chat & 11 & 13 & 22 & 9 & 24 & 9 \\
    Gemini-2.5-Flash & 11 & 14 & 24 & 7 & 25 & 5 \\
    Gemma-2-9B & 12 & 13 & 12 & 12 & 11 & 10 \\
    Gemma-3-12B & 13 & 23 & 15 & 17 & 22 & 14 \\
    LLaMA-3.1-8B & 7 & 18 & 12 & 11 & 22 & 10 \\
    Mistral-7B & 13 & 21 & 17 & 13 & 21 & 9 \\
    GPT-5.1 & 12 & 8 & 12 & 2 & 11 & 5 \\
    GPT-5.2 & 8 & 11 & 10 & 2 & 14 & 3 \\
    GPT-OSS-20B & 8 & 6 & 15 & 6 & 11 & 4 \\
    Qwen-2.5-7B & 4 & 22 & 13 & 16 & 23 & 8 \\
    \hline
  \end{tabular}
  \caption{\textbf{Paired disagreement counts used in McNemar tests.}
  Asymmetric counts indicate representation-specific error patterns.}
  \label{tab:mcnemar_counts}
\end{table}

\end{document}